\documentclass[10pt,twocolumn,letterpaper]{article}

\usepackage{cvpr}
\usepackage{times}
\usepackage{epsfig}
\usepackage{graphicx}
\usepackage{amsmath}
\usepackage{amssymb}

\usepackage[pagebackref=true,breaklinks=true,letterpaper=true,colorlinks,bookmarks=false]{hyperref}

\cvprfinalcopy 

\def\cvprPaperID{1835} 

\ifcvprfinal\fi
\begin{document}

\title{Multi-task Collaborative  Network for Joint  Referring Expression Comprehension and Segmentation}

\author{
	Gen Luo $^{1*}$,
	Yiyi Zhou$^{1}$\thanks{Equal Contribution. $\dagger$ Corresponding Author.},
	Xiaoshuai Sun$^{1}$,
	Liujuan Cao$^{1}$,
	Chenglin Wu$^{2}$,
	Cheng Deng$^{3}$,
	Rongrong Ji$^{1\dagger}$
	 \\
	$^1$Media Analytics and Computing Lab, Department of Artificial Intelligence,\\
	School of Informatics, Xiamen University, 361005, China.\\
	$^2$DeepWisdom, China.
	$^3$Xidian University, China.
	   \\	
	{\tt\small \{luogen, zhouyiyi\}@stu.xmu.edu.cn, \{xssun,caoliujuan\}@xmu.edu.cn},\\
	{\tt\small alexanderwu@fuzhi.ai,chdeng.xd@gmail.com, rrji@xmu.edu.cn}\\
}

\maketitle
\thispagestyle{empty}

\begin{abstract}
	Referring expression comprehension (REC) and segmentation (RES) are two highly-related tasks, which both aim at identifying the referent according to a natural language expression. 
	In this paper, we propose a novel Multi-task Collaborative Network (MCN)\footnote{Source codes and pretrained backbone are available at : \url{https://github.com/luogen1996/MCN}} to achieve a joint learning of REC and RES for the first time.
	In MCN,  RES can help REC to achieve better language-vision alignment, while REC can help RES to  better locate the referent.
	In addition, we address a key challenge in this  multi-task setup, i.e., the prediction conflict, with two innovative designs namely,   Consistency Energy Maximization (CEM) and Adaptive Soft Non-Located Suppression (ASNLS). 
	Specifically, CEM enables  REC and RES  to focus  on similar visual regions by maximizing the consistency energy between  two tasks.  ASNLS  supresses the response of unrelated regions in RES based on the prediction of REC. 
	To validate our model, we conduct extensive experiments on three benchmark datasets of REC and RES, i.e., RefCOCO, RefCOCO+ and RefCOCOg. 
	The experimental results report   the significant  performance gains of MCN over all  existing methods, i.e., up to +7.13\% for REC and +11.50\% for RES over SOTA, which  well confirm the validity of our model  for  joint REC and RES learning.
\end{abstract}

\section{Introduction}

\begin{figure}[t]
	\centering
	\includegraphics[width=1\columnwidth,height=0.65\columnwidth]{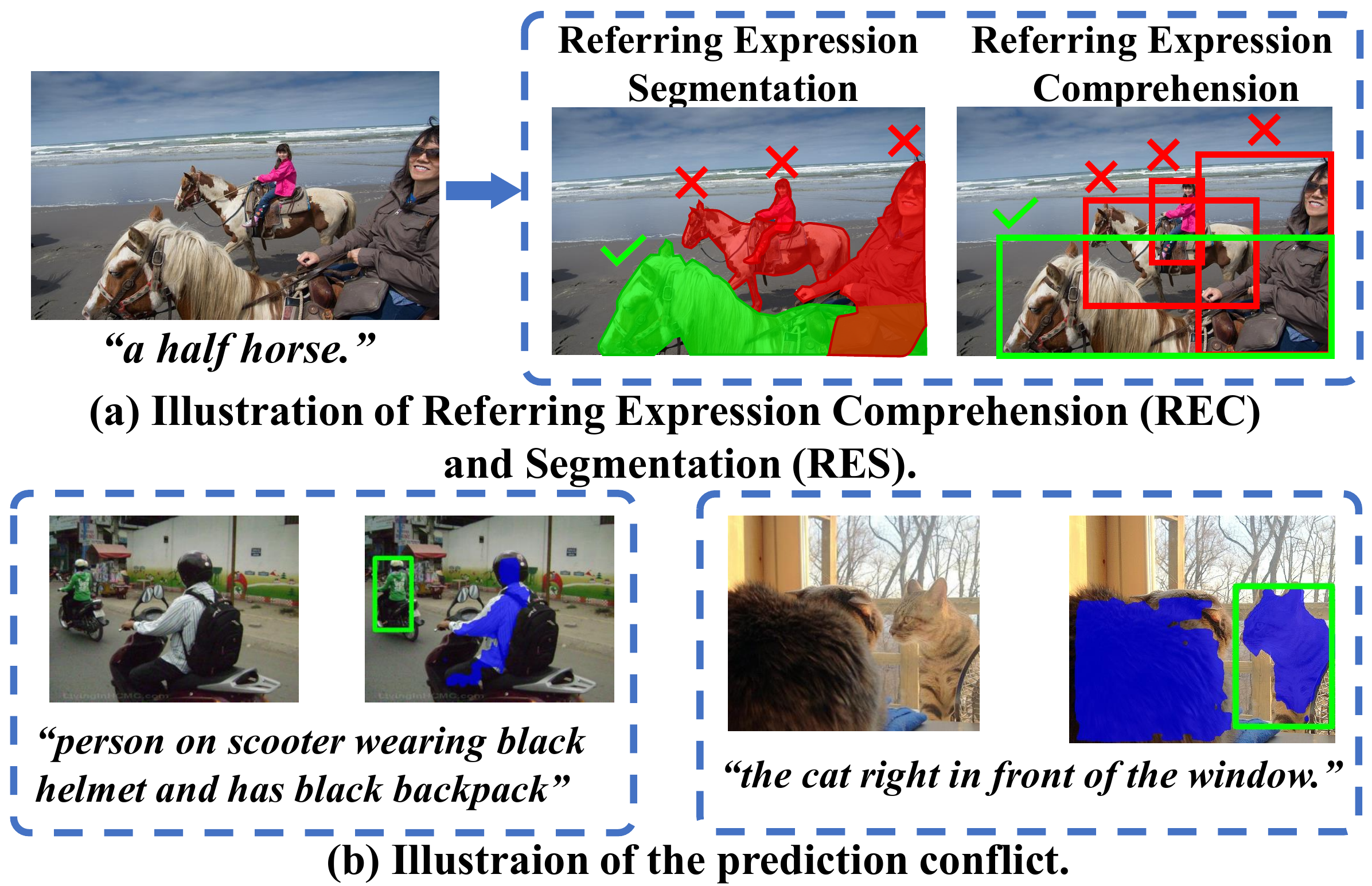}
	\caption{(a) The RES and REC models first perceive the instances in an image and then locate one or few referents based on an expression.  (b)  Two typical cases of prediction conflict: wrong REC correct RES (left) and wrong RES correct REC (right).}
	\label{fig1} 
	\vspace{-1em}
\end{figure}

Referring Expression Comprehension (REC)~\cite{hu2017modeling,hu2016natural,liu2017referring,luo2017comprehension-guided,yu2016modeling,yu2017a,zhang2017discriminative,yu2018mattnet:,wang2019neighbourhood} and Referring Expression Segmentation (RES)~\cite{LSTM-CNN,RRN,CMSA,DMN,KWA} are two   emerging  tasks, which involves identifying  the target  visual  instances according to a given  linguistic  expression. 
Their difference is  that in REC, the targets are grounded by bounding boxes, while   they are segmented in RES,  as shown in Fig.~\ref{fig1}(a).  

REC and RES are  regarded as two seperated tasks  with distinct  methodologies  in the existing literature.
In REC, most existing methods~\cite{hu2017modeling,hu2016natural,liu2017referring,luo2017comprehension-guided,mao2016generation,yu2016modeling,yu2017a,yu2018rethinking,zhang2017discriminative} follow a multi-stage pipeline, \emph{i.e.,}  detecting the salient regions from the image and selecting the most matched one through multimodal interactions.
In RES, existing methods~\cite{LSTM-CNN,RRN} usually embed a language module, \emph{e.g.}, LSTM or GRU~\cite{GRU}, into a  one-stage segmentation network like FCN~\cite{FCN} to segment the referent. 
Although some recent works like MAttNet~\cite{MATT:} can  simultaneously process  both REC and RES,  their multi-task functionality are largely attributed to their backbone detector, \emph{i.e.}, MaskRCNN~\cite{MATT:}, rather than explicitly interacting and reincecing  two tasks.

It is a natural thought to jointly learn REC and RES to reinforce 
each other, as similar to the classic endeavors in joint object detection and segmentation~\cite{fu2019retinamask,he2017mask,dvornik2017blitznet}.
Compared with RES, REC is superior in predicting the potential location of the referent, which can compensate for the deficiency of RES in determining  the correct instance. 
On the other hand, RES is trained with pixel-level labels, which can  help REC obtain better language-vision alignments during the multimodal training. However, such a joint learning is not trivial at all. We attribute the main difficulty to \emph{the prediction conflict}, as shown in Fig.~\ref{fig1}~(b). Such  prediction conflict is also common in general detection and segmentation based multi-task models~\cite{he2017mask,eigen2015predicting,chen2018driving}. However, it is more prominent in RES and REC, since only one or a few of the multiple instances are the correct referents.

To this end, we propose a novel Multi-task Collaborative Network (MCN) to jointly learn REC and RES in a one-stage fashion,  which is illustrated  in Fig.~\ref{fig2}. 
The principle of  MCN is a multimodal and multitask collaborative learning framework.  It   links  two tasks centered on the language information  to maximize their collaborative learning. 
Particularly, the visual backbone and the language encoder are shared, while the multimodal inference branches of two tasks  remain relatively separated.
Such a design is to take full account of the intrinsic differences between REC and RES,  and avoid  the performance  degeneration of one task   to accommodate the other, \emph{e.g.},  RES typically requires higher resolution feature maps for its pixel-wise prediction.

To address the issue  of prediction conflict, we equip MCN with two innovative designs, namely \emph{Consistency Energy Maximization} (CEM) and \emph{Adaptive Soft Non-Located Suppression} (ASNLS).  
CEM is a language-centric loss function that  forces two tasks on  the similar visual areas by maximizing the consistency energy between two inference branches. Besides, it also serves as a pivot to connect the learning processes of REC and RES.  ASNLS is a  post-processing method, which  suppresses the response of unrelated regions in RES based on the prediction of REC. Compared with existing hard processing methods, \emph{e.g.,} RoI-Pooling~\cite{ren2017faster} or Rol-Align~\cite{he2017mask}, the adaptive  soft processing of ASNLS  allows the model to have a higher error tolerance in terms of the detection results. 
With CEM and ASNLS, MCN can significantly reduce the effect of the prediction conflict, as validated in our quantitative evaluations.

To validate our approach, we conduct extensive experiments on three benchmark datasets, \emph{i.e.}, RefCOCO, RefCOCO+ and RefCOCOg, and compare MCN to a set of state-of-the-arts (SOTAs) in both REC and RES~\cite{yu2018mattnet:,yang2019fast,CMSA,RRN,liu2019learning,wang2019neighbourhood}.
Besides, we propose a new metric termed  \emph{Inconsistency Error} (IE) to objectively measure the impact of \emph{prediction conflict}. 
The experiments show  superior performance gains of MCN over  SOTA, \emph{i.e.},  up to +7.13\%  in REC and +11.50\%  in RES. 
More importantly, these experimental results greatly validate  our argument of reinforcing REC and RES in a joint framework, and the impact of prediction conflict is  effectively reduced by our   designs. 

Conclusively, our contributions are  three-fold:
\begin{itemize}
	\item  We propose a new multi-task network for REC and RES, termed  Multi-task Collaborative Network (MCN), which facilitates  the collaborative learning  of REC and RES. 
	
	\item We address the key issue in the collaborative learning of REC and RES, \emph{i.e.}, the prediction conflict, with two innovative designs, \emph{i.e.},  \emph{Consistency Energy Maximization} (CEM) and \emph{Adaptive Soft Non-Located Suppression} (ASNLS)
	
	\item The proposed MCN has established  new state-of-the-art performance  in both REC and RES on three benchmark datasets, \emph{i.e.}, RefCOCO, RefCOCO+ and RefCOCOg. 
	Notably, its inference speed is 6 times faster than that of  most existing multi-stage methods in REC. 
\end{itemize}

\section{Related Work}

\subsection{Referring Expression Comprehension}
Referring expression comprehension (REC) is a task of grounding  the target object with a bounding box based on a given  expression. Most existing methods~\cite{hu2017modeling,hu2016natural,liu2017referring,luo2017comprehension-guided,yu2016modeling,yu2017a,zhang2017discriminative,yu2018mattnet:,wang2019neighbourhood} in REC follow a multi-stage procedure  to select  the best-matching region from a set of candidates. Concretely, a pre-trained detection network, \emph{e.g.}, FasterRCNN~\cite{ren2017faster}, is first  used to detect salient regions of a given image. 
Then to rank the query-region pairs,  a multimodal embedding network~\cite{rohrbach2016grounding,wang2016learning,liu2017referring,chen2017query-guided,zhang2018grounding}  is used,  or  the visual features  are included  into the language modeling~\cite{mao2016generation,baxter2000model,luo2017comprehension-guided,hu2016natural,yu2016modeling}. 
Besides, additional processes are also used to improve  the multi-modal ranking  results, \emph{e.g.}, the prediction of image attributes~\cite{MATT:} or the calculation of location features~\cite{yu2017a,wang2019neighbourhood}. 
Despite their high performance, these methods  have a significant drawback  in low  computational efficiency. Meanwhile, their  upper-bounds are largely determined by the pre-trained object detector~\cite{sadhu2017zero}. 

To speedup the inference, some recent works in REC resort to a  one-stage modeling~\cite{sadhu2017zero,yang2019fast}, which embeds the extracted linguistic feature into a one-stage detection network, \emph{e.g.,}  YoloV3~\cite{redmon2018yolov3:}, and directly predicts the bounding box. However, their performance is still worse than the most popular two-stage approaches, \emph{e.g.,} MattNet~\cite{yu2018mattnet:}. Conclusively, our work are the first  to combine REC and RES in a one-stage framework, which not only boosts the inference speed but also outperforms  these two-stage methods.

\begin{figure*}[t]
	\centering
	\includegraphics[width=2\columnwidth,height=0.55\columnwidth]{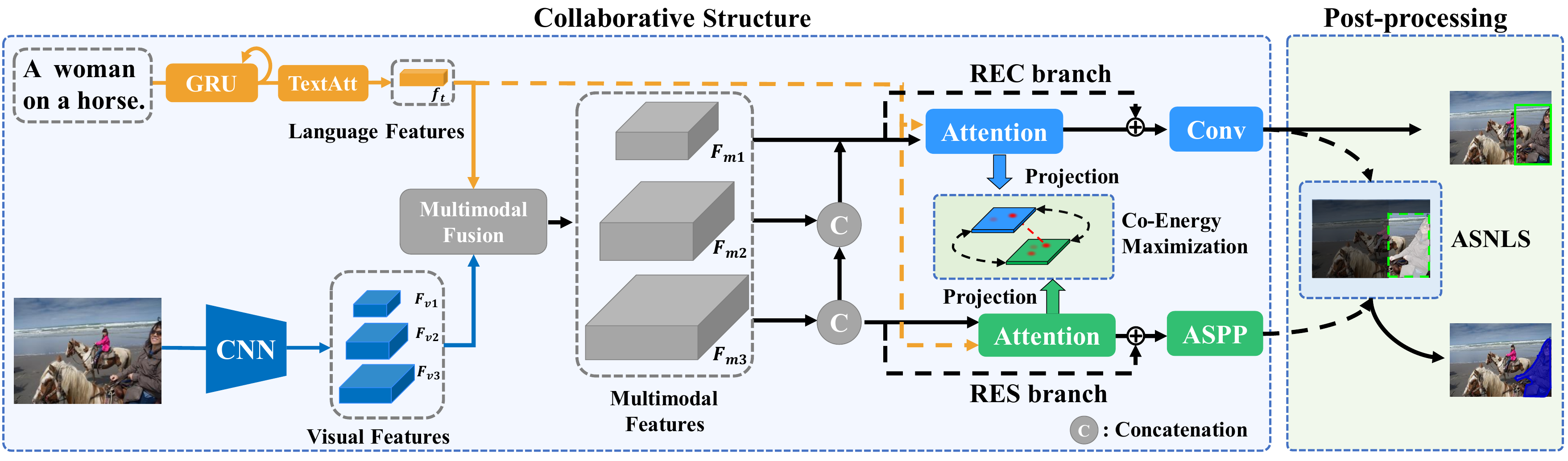}
	\caption{The framework of the proposed \emph{Multi-task Collaborative Network} (MCN). The visual features and linguistic  features are extracted by a deep convolutional network and a bi-GRU network respectively, and then fused to generate the  multi-scale multimodal features. The bottom-up connection from the RES branch effectively promotes the language-vision alignment of REC.  The two branches   are further reinforced by each other through  CEM. Finally, the  output of RES is  adaptively refined by ASNLS based on the REC result.}
	\label{fig2} 
	\vspace{-0.5em}
\end{figure*}
\subsection{Referring Expression Segmentation}
Referring expression segmentation (RES) is a task of segmenting  the  referent according to a given textual expression. A typical solution of RES is to embed the language encoder into a  segmentation network, \emph{e.g.}, FCN~\cite{FCN}, which further learns a multimodal tensor for decoding the segmentation mask~\cite{LSTM-CNN,RRN,DMN,CMSA,KWA}. Some recent developments also focus on improving the efficiency of multimodal interactions, \emph{e.g.},  adaptive feature fusions at multi-scale~\cite{LSTM-CNN},  pyramidal fusions for progressive refinements~\cite{RRN,DMN}, and query-based or transformer-based attention modules~\cite{KWA,CMSA}. 

Although  relatively high performance is achieved in RES, existing methods are generally inferior in determining the referent compared to  REC. To explain, the pixel-wise prediction of RES  is easy to generate uncertain segmentation mask that includes incorrect regions or objects, \emph{e.g.}, overlapping people. 
In this case, the incorporation of REC can help  RES to suppress responses of unrelated regions, while activating the related ones based on the predicted bounding boxes.

\subsection{Multi-task Learning}
Multi-task Learning (MTL) is often applied when related tasks can be performed simultaneously. 
MTL has been widely deployed in a variety of computer vision tasks~\cite{eigen2015predicting,chen2018driving,nekrasov2019real,dvornik2017blitznet,he2017mask,ubernet}. 
Early endeavors~\cite{eigen2015predicting,chen2018driving,nekrasov2019real}  resort to learn multiple  tasks of pixel-wise predictions in an MTL setting, such as depth estimation, surface normals or semantic segmentation. 
Some recent works also focus on combining the object detection and segmentation into a joint framework, \emph{e.g.}, MaskRCNN~\cite{he2017mask}, YOLACT~\cite{yolact}, and RetinaMask~\cite{fu2019retinamask}.
The main difference between MCN and these methods is that MCN is an MTL network centered on the language information. 
The selection of target instance in REC and RES also exacerbates the issue of prediction conflicts, as mentioned above.

\section{Multi-task Collaborative Network}
The framework of the proposed \emph{Multi-task Collaborative Network} (MCN) is shown in Fig.~\ref{fig2}. Specifically, the representations of the input image and expression are first extracted by the visual and the language encoders respectively, which are further fused to obtain the multimodal features of different scales. 
These multimodal features are then fed to the inference branches of REC and RES, where a bottom-up connection is built to strengthen the collaborative learning of two tasks. 
In addition, a language-centric connection is also built between two branches, where the \emph{Consistency Energy Maximization} loss is used to maximize the consistency energy between REC and RES.  
After inference, the proposed  
\emph{Adaptive Soft Non-Located Suppression} (ASNLS) is used to refine the segmentation result of RES based on the predicted bounding box by the REC branch.

\subsection{The Framework}\label{framework}
As shown in Fig.~\ref{fig2}, MCN is partially shared, where the inference branches of RES and REC remain relatively independent.
The intuition is two-fold: 
On one hand, the objectives of two tasks are still distinct, thus the full sharing of the inference branch can be counterproductive. On the other hand, such a relatively independent design  enables the optimal settings of two tasks, \emph{e.g.}, the resolution of feature map.

Concretely, given an image-expression pair $\left(I,E\right) $,  we first use the visual backbone to  extract the  feature maps of three scales, denoted as $\mathbf{F}_{v_1} \in \mathbb{R}^{{h_1} \times {w_1}\times   d_1}, \mathbf{F}_{v_2} \in \mathbb{R}^{{h_2}\times {w_2}\times   d_2}, \mathbf{F}_{v_3} \in \mathbb{R}^{{h_3}\times {w_3}\times   d_3}$,  where $h$, $w$ and $d$ denote the height, width and the depth.  The expression is processed by a bi-GRU encoder, where the  hidden states are weightly combined as the textual feature by using a self-guided attention module~\cite{yang2016hierarchical}, denoted as $f_t\in \mathbb{R}^{d_t}$.  

Afterwards, we   obtain the first multimodal tensor by fusing $\mathbf{F}_{v_1}$ with $f_t$, which is formulated as:
\begin{equation}
f_{m_1}^l=\sigma(f_{v_1}^l\mathbf{W}_{v_1}) \odot \sigma(f_t\mathbf{W}_t),
\label{eq1}
\end{equation}
where  $\mathbf{W}_{v_1}$ and $\mathbf{W}_{t}$ are the projection weight matrices, and $\sigma$ denotes  \emph{Leaky ReLU}~\cite{LRELU}.
$f_{m_1}^l$ and $f_{v_1}^l$ are the feature vector of $\mathbf{F}_{m_1}$ and $\mathbf{F}_{v_1}$, respectively. Then, the other two multimodal tensors, $\mathbf{F}_{m_2}$ and $\mathbf{F}_{m_3}$, are obtained by the following procedure: 
\begin{equation}
\begin{aligned}
&\mathbf{F}_{m_{i-1}}=UpSample(\mathbf{F}_{m_{i-1}}), \\
&\mathbf{F}_{m_{i}}=[\sigma(\mathbf{F}_{m_{i-1}}\mathbf{W}_{m_{i-1}}),\sigma( \mathbf{F}_{v_{i}}\mathbf{W}_{v_{i}})],
\end{aligned}
\label{eq2}
\end{equation}
where $i\in\{2,3\}$, \emph{UpSampling} has a stride of $2\times 2$,   and [$\cdot$] denotes concatenation. 

Such a multi-scale fusion   not only propagates the language information through upsamplings and concatenations, but also  includes the mid-level semantics to the upper feature maps, which is crucial for both REC and RES. Considering that these two tasks have different requirements for the feature map scales, \emph{e.g.}, $13\times13$ for REC  and $52\times 52$ for RES, we use  $\mathbf{F}_{m_1}$  and $\mathbf{F}_{m_3}$ as the inputs of REC and RES, respectively.

To further strengthen the connection of two tasks, we  implement another bottom-up path from RES to REC. 
Such a connection   introduces the semantics supervised by the pixel-level labels in RES to benefit the language-vision alignments in REC. Particularly, the new multimodal tensor,  $\mathbf{F}_{m_1}'$  for REC,  is obtained by repeating the down sampling and concatenations twice, as  similar to the procedure defined in Eq.~\ref{eq2}.  Afterwards,  $\mathbf{F}_{m_1}'$ and $\mathbf{F}_{m_3}'$ for REC and RES   respectively are then refined by two \emph{GARAN Attention} modules~\cite{zhou2019GIN}, as illustrated in Fig.~\ref{fig2}.

\textbf{Objective Functions.}
For RES,  we implement the ASPP decoder~\cite{DEEPLAB} to predict the segmentation mask based on the refined multimodal tensor. Its loss function is defined by
\begin{equation}
\ell_{res}=-\sum_{l=1}^{h_3 \times w_3}\big[g_l {\log}\left(o_l\right)+\left(1-g_l\right)\log\left(1-o_l\right)\big], 
\end{equation}
where $g_l$ and $o_l$ represent the elements of the down-sampled   ground-truth $\mathbf{G'} \in \mathbb{R}^{52 \times 52}$ and  predicted mask $\mathbf{O} \in \mathbb{R}^{52 \times 52}$,  respectively.

For REC, we add a regression layer after the multimodal tensor for predicting the confidence score and the bounding box of the referent. 
Following the setting in YoloV3~\cite{redmon2018yolov3:}, the regression loss of REC is formulated as:
\begin{equation}
\ell_{rec}=\sum_{l=1}^{h_1 \times w_1 \times N} \ell_{box}\left(t_l^*,t_l\right)+\ell_{conf}\left(p_l^*,p_l\right), 
\label{6}
\end{equation}
where $t_l$ and  $p_l$ are the predicted  coordinate position of the  box and confidence score. $N$ is the number of anchors for each grid.  $t_l*$ and $p_l*$ are the ground-truths.   $p_l^*$ is set to 1 when the anchor  matches  ground-truth.  $\ell_{box}$ is a  binary cross-entropy to measure the regression loss for the center point of the bounding box. For the width and height of the bounding box, we adopt the smooth-L1 loss~\cite{ren2017faster}.  $\ell_{conf}$ is the binary cross entropy.

\subsection{Consistency Energy Maximization}
\begin{figure}[t]
	\centering
	\includegraphics[width=0.85\columnwidth,height=0.5\columnwidth]{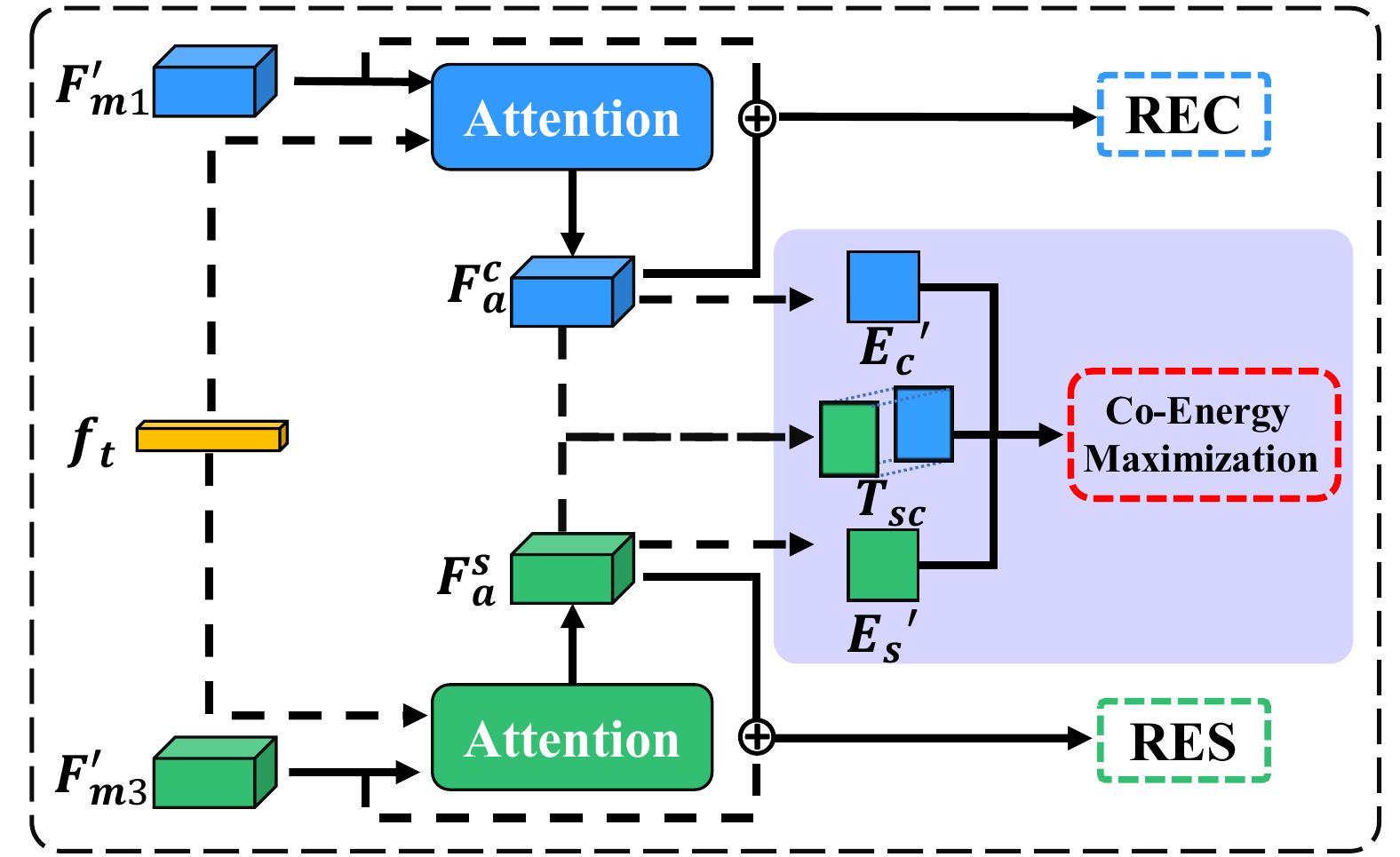}
	\caption{Illustration of the Consistency Energy Maximization (CEM). The CEM loss optimizes the attention features to maximize the consistency spatial responses between REC and RES. }
	\label{cem} 
	\vspace{-0.6em}
\end{figure}

We further propose a \emph{Consistency Energy Maximization} (CEM) scheme  to theoretically reduce the impact of prediction conflict.
As shown in Fig.~\ref{cem},  CEM build  a language-centered connection between two branches. Then, CEM loss defined in Eq.~\ref{cem_loss} is used to maintain the consistency of  spatial responses  for two tasks by maximizing the energy between their  attention tensors. 

Concretely, given the attention tensors of RES and REC, denoted as $\mathbf{F}_a^s \in \mathbb{R}^{(h_3\times w_3) \times d}$ and $\mathbf{F}_a^c \in \mathbb{R}^{(h_1\times w_1)                                                                                                                                                                                                                                                                                                                                                                                                                                                                                                                                                                                                                                                                                                                                                                                                                                                                                                                                                                                                                                                                                                                                                                                                                                                                                                                                                                                                                                                                                                                                                                                                                                                                                                                                                                                                                                                                                                                                                                                                                                                                                                                                                                                                                                                                                                                                                                                                                                                                                                                                                                                                                                                                                                                                                                                                                                                                                                                                                                                                                                                                                                                                                                                                                                                                                                                                                                                                                                                                                                                                                                                                                                                                                                                                                                                                                                                                                                                                                                                                                                                                                                                                                                                                                                                                                                                                                                                                                                                                                                                                                                                                                                                                                                                                                                                                                                                                                                                                                                                                                                                                                                                                                                                                                                                                                                                                                                                                                                            \times d}$, 
we project them to the two-order tensors by: 
\begin{equation}
\begin{aligned}
&E_s=\mathbf{F}_{a}^s\mathbf{W}_s,&E_c=\mathbf{F}_{a}^c\mathbf{W}_c,
\end{aligned}
\end{equation}
where $\mathbf{W}_s,\mathbf{W}_c \in \mathbb{R}^{d\times1}$, $E_s \in \mathbb{R}^{(h_3\times w_3)}$ and $E_c \in \mathbb{R}^{(h_1\times w_1)}$.  
Afterwards, we perform \emph{Softmax} on $E_c$ and $E_s$ to obtain the energy distributions of REC and RES over the  image, dentoed as $E_c'$ and $E_s'$.
Elements of $E_c'$ and  $E_s'$ indicate the response degrees of the corresponding regions  towards the given expression.  

To maximize the co-energy between two tasks, we further calculate the inter-task correlation, $\mathbf{T}_{sc}\in \mathbb{R}^{(h_3 \times w_3)\times (h_1 \times w_1)}$, by 
\begin{equation}
\begin{aligned}
\mathbf{T}_{sc}(i,j)=s_w*\frac{{f_{i}^s}^T {f_{j}^c}}{{ \| {f_{i}^s}  \|}{  \| {f_{j}^c}  \|}}+s_b, 
\end{aligned}
\end{equation}
where the $f_{i}^s \in \mathbb{R}^{d} $ and $f_{j}^c \in \mathbb{R}^{d}$ are elements of  $\mathbf{F}_{a}^s$ and $\mathbf{F}_{a}^c$, respectively. The $s_w$ and $s_b$ are two scalars to scale the value in $\mathbf{T}_{sc}$ to $\left(0,1 \right]$.
The  co-energy $C$  is caculated as: 
\begin{equation}
\begin{aligned}
C\left(i,j\right)=&  \log \big[E_s'\left(i\right)\mathbf{T}_{sc}\left(i,j\right)E_c'\left(j\right)\big]\\
=& E_s\left(i\right) + E_c\left(j\right) +\log \mathbf{T}_{sc}\left(i,j\right)\\& - \log 
\alpha_s- \log \alpha_c,
\end{aligned}	
\end{equation}
where the $\alpha_s$ and $\alpha_c$ are two reguralization term to penalize  the irrelevant responeses, denoted as:
\begin{equation}
\begin{aligned}
\alpha_s=\sum_{i=1}^{h_3\times w_3}e^{E_s\left(i\right)}, \alpha_c=\sum_{i=1}^{h_1\times w_1}e^{E_c\left(i\right)}.
\end{aligned}	
\end{equation}
Finally, the CEM loss is formulated by 
\begin{equation}
\begin{aligned}
\ell_{cem}=-\sum_{i=1}^{h_3\times w_3}\sum_{j=1}^{h_1 \times w_1}C(i,j).
\end{aligned}	
\label{cem_loss}
\end{equation}
\begin{figure}[t]
	\centering
	\includegraphics[width=1\columnwidth,height=0.63\columnwidth]{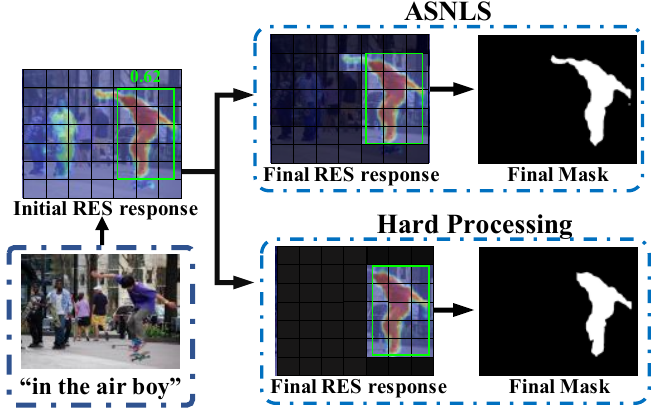}
	\caption{The comparison between ASNLS and conventional hard processing (bottom).  Compared to the hard processing,  ASNLS has a better error tolerance for REC predictions, which can well  preserve the integrity of  referent given an inaccurate box.}
	\label{fig3} 
	\vspace{-2em}
\end{figure}

\subsection{Adaptive Soft Non-Located Suppression}
We further  propose a soft post-processing method to methodically  address the prediction conflict, termed as \emph{Adaptive Soft Non-Located Suppression} (ASNLS).
Based on the predcited bounding box by REC, ASNLS suppresses the response of unrelated regions and strengths the related ones. 
Compared to the existing hard processings, \emph{e.g.}, ROI Pooling~\cite{ren2017faster} and ROI Align~\cite{he2017mask}, which directly crop features of the  bounding box, the soft processing of ASNLS can obtain a better error tolerance towards the predictions of REC, as illustrated in Fig.~\ref{fig3}. 

In particular,  given the predicted mask by the RES branch, $\mathbf{O} \in \mathbb{R}^{h_3 \times w_3}$, and the bounding box $b$, each element $o_i$ in $\mathbf{O}$ is updated by:
\begin{equation}
m_i=\left\{\begin{matrix}
\alpha_{up} * o_i, & if~o_i~in~ b, \\ 
\alpha_{dec} * o_i, & else.
\end{matrix}\right.
\label{eq11}
\end{equation}
Here, $\alpha_{up} \in \left( 1,+\infty \right)$ and $\alpha_{dec} \in \left( 0,1 \right) $ are the enhancement and  decay factors, respectively. We term this method  in~Eq.~\ref{eq11} as \emph{Soft Non-Located Suppression} (Soft-NLS).   After that, the updated RES result  $\mathbf{O}$ is binarized by  a threshold to generate the final mask.

In addition, we extend the Soft-NLS to an adaptive version, where the update factors are determined by the prediction confidence of REC. To explain, a lower confidence $p$  indicates  a larger uncertainty that  the referent can be segmented integrally, and should increase the effects of NLS to eliminate the uncertainty as well as to enhance its saliency.  Specifically, given the confidence score $p$, $\alpha_{up}$ and $\alpha_{dec}$ are calculated by
\begin{equation}
\begin{aligned}
&\alpha_{up}=\lambda_{au}*p+\lambda_{bu},\\
&\alpha_{dec}=\lambda_{ad}*p+\lambda_{bd},
\end{aligned}
\label{eq10}
\end{equation}
where  the $\lambda_{au}$,  $\lambda_{ad}$ ,  $\lambda_{bu}$ and $\lambda_{bd}$  are hyper-parameters\footnote{In our experiments, we set $\lambda_{au}=-1$,  $\lambda_{ad}=1$,  $\lambda_{bu}=2$,  $\lambda_{bd}=0$.} to control  the  enhancement and decay, respectively. We term this adaptive approach as \emph{Adaptive Soft Non-Located Suppression} (ASNLS).   

\begin{table*}[t]
	\centering
	\caption{Comparisons of the different post-processing methods  on the validation set of RefCOCO. $\downarrow$ denotes the lower is better. }
	\begin{tabular}{|l|c|c|c|c|c|c|c|}
		\hline
		\multicolumn{1}{|c|}{} & IoU & Acc@0.5 & Acc@0.6 & Acc@0.7 & Acc@0.8 & Acc@0.9 & IE~$\downarrow$ \\ \hline
		w.o. post-processing & 61.61 & 73.95 & 67.42 & 56.39 & 32.02 & 4.72 & 10.37\% \\
		RoI Crop~\cite{he2017mask,ren2017faster} & 61.19 & 75.13 & 68.88 & 57.61 & 32.42 & 3.81 & 7.91\% \\
		Soft-NLS~(ours) & 62.27 & 75.92 & 69.48 & 58.21 & 33.20 & 5.11 & 7.28\% \\
		ASNLS~(ours) & 62.44 & 76.60 & 70.33 & 58.39 & 33.68 & 5.26 & 6.65\% \\ \hline
	\end{tabular}
	
	\label{tab2}
	\vspace{-0.5em}
\end{table*}

\begin{table*}[t]
	\centering
	\caption{Ablation study on the \emph{val} set of three datasets. The metric is Acc@0.5 for REC, and IoU for RES. \emph{Base} indicates the  network structure  without any extra components.}
	
	\begin{tabular}{|l|c|c|c|c|c|c|c|l|l|}
		\hline
		& \multicolumn{3}{c|}{RefCOCO} & \multicolumn{3}{c|}{RefCOCO+} & \multicolumn{3}{c|}{RefCOCOg} \\ \hline
		\multicolumn{1}{|c|}{} & REC & RES & IE~$\downarrow$  & REC & RES & IE~$\downarrow$  & REC & \multicolumn{1}{c|}{RES} & \multicolumn{1}{c|}{IE~$\downarrow$} \\ \hline
		MCN~(Base) & 77.45 & 58.24 & 13.80\% & 62.74 & 44.08 & 20.70\% & 62.29 & 44.58 & 19.87\% \\ \hline
		+TextAtt & \multicolumn{1}{l|}{77.65} & \multicolumn{1}{l|}{58.44} & \multicolumn{1}{l|}{13.44\%} & \multicolumn{1}{l|}{63.07} & \multicolumn{1}{l|}{44.38} & \multicolumn{1}{l|}{19.88\%} & \multicolumn{1}{l|}{64.51} & 46.58 & 18.71\% \\ \hline
		+GARAN & 79.20 & 59.07 & 13.37\% & 66.22 & 47.89 & 17.12\% & 65.98 & 47.33 & 17.44\% \\ \hline
		+CEM & 80.08 & 61.61 & 10.37\% & 67.16 & 49.55 & 13.51\% & 66.46 & 48.56 & 14.90\% \\ \hline
		+ASNLS & 80.08 & 62.44 & 6.65\% & 67.16 & 50.62 & 7.54\% & 66.46 & 49.22 & \multicolumn{1}{c|}{9.41\%}\\ \hline
	\end{tabular}
	\label{tab3}
	\vspace{-1em}
\end{table*}
\begin{table}[t]
	\centering
	\caption{Comparisons of MCN with different  network structures on the \emph{val} set of RefCOCO. The structure of MCN can significantly improve the performance of both two tasks, and it is also superior than other single and multi-task frameworks.}
	\resizebox{80mm}{16.8mm}{
		\setlength{\tabcolsep}{1mm}{
			\begin{tabular}{|l|c|c|}
				\hline
				\multicolumn{1}{|c|}{\textbf{Structure}} & REC & RES \\ \hline
				Single\_REC(scale\footnotemark[1]=$13^2$) & 70.38 & - \\
				Single\_REC(scale=$52^2$) & 68.58 & - \\
				Single\_RES(scale=$13^2$) & - & 36.37 \\
				Single\_RES(scale=$52^2$) & - & 57.91 \\ \hline
				OnlyHeadDifferent(scale=$13^2$) & 72.42 & 34.50 \\
				OnlyHeadDifferent(scale=$52^2$) & 72.54 & 58.08 \\
				OnlyBackboneShared(REC\_scale=$13^2$, RES\_scale=$52^2$) & 75.81 & 58.16 \\ \hline
				\textbf{MCN~(Base)} & \textbf{77.45} & \textbf{58.24} \\ \hline
			\end{tabular}
	}}
	\label{tab1}
	\vspace{-1em}
\end{table}

\subsection{Overall Loss}

The overall loss function of MCN is  formulated as:
\begin{equation}
\ell_{all}=\lambda_{s}\ell_{res}+\lambda_{c}\ell_{rec}+\lambda_{e}\ell_{cem},
\end{equation}
where, $\lambda_{s}$, $\lambda_{c}$ and $\lambda_{e}$ control the relative importance among the three losses, which are set to 0.1,  1.0 and 1.0 in our experiments, respectively.

\section{Experiments}
We further evaluate the proposed  MCN
on three benchmark datasets, i.e., RefCOCO~\cite{REFCOCO},
RefCOCO+~\cite{REFCOCO} and RefCOCOg~\cite{REFCOCOG}, and compare them
to a set of state-of-the-art methods~\cite{MATT:,wang2019neighbourhood,yang2019fast,CMSA,RRN} of both REC and RES.

\subsection{Datasets}
\textbf{RefCOCO}~\cite{REFCOCO} has 142,210 referring
expressions for 50,000 bounding boxes in 19,994 images
from MS-COCO~\cite{MSCOCO}, which is split into \emph{train}, \emph{validation},
\emph{Test A} and \emph{Test B} with a number of 120,624, 10,834, 5,657
and 5,095 samples, respectively. The expressions are collected
via an interactive game interface~\cite{REFCOCO}, which are
typically short sentences with a average length of 3.5 words.
The categories of bounding boxes in TestA are people while
the ones in TestB are objects.

\textbf{RefCOCO+}~\cite{REFCOCO} has 141,564 expressions for 49,856
boxes in 19,992 images from MS-COCO. It is also divided
into splits of \emph{train} (120,191), \emph{val} (10,758), \emph{Test A} (5,726)
and \emph{Test B} (4,889). Compared to RefCOCO, its expressions
include more appearances (attributes) than absolute
locations. Similar to RefCOCO, expressions of Test A in RefCOCO+ are about people
while the ones in Test B are about objects.

\textbf{RefCOCOg}~\cite{REFCOCOG,nagaraja2016modeling} has 104,560 expressions for 54,822
objects in 26,711 images. In this paper, we use the UNC partition~\cite{nagaraja2016modeling}  for training and testing our method. 
Compared to RefCOCO and RefCOCO+, expressions in
RefCOCOg are collected in a non-interactive way, and the
lengths are longer (8.4 words on average), of which content
includes both appearances and locations of the referent.

\begin{table*}[t]
	\centering
	\caption{Comparisons of MCN with the state-of-the-arts on the REC task. }
	{
		\begin{tabular}{|l|l|c|c|c|c|c|c|c|c|c|}
			\hline
			\multicolumn{1}{|c|}{}      & \multicolumn{1}{c|}{}                & \multicolumn{3}{c|}{RefCOCO}                     & \multicolumn{3}{c|}{RefCOCO+}                    & \multicolumn{2}{c|}{RefCOCOg}   & \multicolumn{1}{l|}{} \\ \cline{3-10}
			\multicolumn{1}{|c|}{Model} & \multicolumn{1}{c|}{Visual Features} & val            & testA          & testB          & val            & testA          & testB          & val            & test           & Speed*~$\downarrow$            \\ \hline
			MMI~\cite{mao2016generation}~\tiny{CVPR16}                         & vgg16                                & -              & 64.90           & 54.51          & -              & 54.03          & 42.81          & -              & -              & \multicolumn{1}{c|}{-} \\
			CMN~\cite{rohrbach2016grounding}~\tiny{CVPR16}                         & vgg16                                & -              & 71.03          & 65.77          & -              & 54.32          & 47.76          & -              & -              & \multicolumn{1}{c|}{-} \\
			\textbf{Spe}+Lis+Rl~\cite{yu2017a}~\tiny{CVPR17}                  & frcnn-resnet101                      & 69.48          & 73.71          & 64.96          & 55.71          & 60.74          & 48.80          & 60.21          & 59.63          &              -         \\
			Spe+\textbf{Lis}+Rl~\cite{yu2017a} ~\tiny{CVPR17}                  & frcnn-resnet101                      & 68.95          & 73.10          & 64.85          & 54.89          & 60.04          & 49.56          & 59.33          & 59.21          &           -            \\
			ParalAttn~\cite{zhuang2018parallel}~\tiny{CVPR18}                & frcnn-vgg16                          & -              & 75.31          & 65.52          & -              & 61.34          & 50.86          & -              & -              &     -                  \\
			LGRANs~\cite{wang2019neighbourhood}~\tiny{CVPR19}                      & frcnn-vgg16                          & -              & 76.60          & 66.40          & -              & 64.00          & 53.40          & -              & -              &      -                 \\
			NMTree~\cite{liu2019learning}~\tiny{ICCV19} & frcnn-vgg16 &71.65& 74.81 &67.34& 58.00 &61.09& 53.45  &61.01& 61.46&-\\
			FAOA~\cite{yang2019fast}~\tiny{ICCV19}               & darknet53                            & 71.15                      & 74.88                      & 66.32                      &    56.86                        &    61.89                        &  49.46
			&  59.44                                   &  58.90                          &  \multicolumn{1}{c|}{
				\textbf{\underline{39 ms}}} \\
			
			MattNet~\cite{MATT:}~\tiny{CVPR18}                        & frcnn-resnet101                      & \multicolumn{1}{l|}{76.40} & \multicolumn{1}{l|}{80.43} & \multicolumn{1}{l|}{69.28} & \multicolumn{1}{l|}{64.93} & \multicolumn{1}{l|}{70.26} & \multicolumn{1}{l|}{56.00} & \multicolumn{1}{l|}{\textbf{\underline{66.67}}} & \multicolumn{1}{l|}{67.01} & \multicolumn{1}{c|}{367 ms} \\
			
			MattNet~\cite{MATT:}~\tiny{CVPR18}                      & mrcnn-resnet101                      & \underline{76.65}          & \underline{81.14}          & \underline{69.99}          & \underline{65.33}          & \underline{71.62}          & \underline{56.02}          & {66.58} & \textbf{\underline{67.27}} &  378 ms                     \\ \hline
			MCN~(ours)                        & vgg16                                & 75.98          & 76.97          & 73.09          & 62.80           & 65.24          & 54.26          &62.42                & 62.29               &48 ms                     \\
			
			MCN~(ours)  &darknet53&\textbf{80.08}&	\textbf{82.29}&	\textbf{74.98}&	\textbf{67.16}&	\textbf{72.86}&	\textbf{57.31}
			&66.46&66.01& 56 ms\\ \hline
		\end{tabular}
	}
	\label{tab4}
	\leftline{
		* The inference time is tested on the same hardware, i.e., GTX1080ti. }
	\vspace{-1.5em}
\end{table*}
\begin{table*}[t]
	\centering
	\caption{Comparisons of MCN with the state-of-the-arts on the RES task.}
	{
		\begin{tabular}{|l|l|c|c|c|c|c|c|c|c|}
			\hline
			\multicolumn{1}{|c|}{}      & \multicolumn{1}{c|}{}                & \multicolumn{3}{c|}{RefCOCO}                                                         & \multicolumn{3}{c|}{RefCOCO+}                                                        & \multicolumn{2}{c|}{RefCOCOg} \\ \cline{3-10} 
			\multicolumn{1}{|c|}{Model} & \multicolumn{1}{c|}{Visual Features} & val                        & testA                      & testB                      & val                        & testA                      & testB                      & val           & test          \\ \hline
			DMN~\cite{DMN}~\tiny{ECCV18}                        & resnet101                            & \multicolumn{1}{l|}{49.78} & \multicolumn{1}{l|}{54.83} & \multicolumn{1}{l|}{45.13} & \multicolumn{1}{l|}{38.88} & \multicolumn{1}{l|}{44.22} & \multicolumn{1}{l|}{32.29} & -             & -             \\
			RRN~\cite{RRN}~\tiny{CVPR18}                         & resnet101                            & 55.33                      & 57.26                      & 53.93                      & 39.75                      & 42.15                      & 36.11                      & -             & -             \\
			CMSA~\cite{CMSA}~\tiny{CVPR19}                        & resnet101                            & \underline{58.32}                      & 60.61                      & \underline{55.09}                      & 43.76                      & 47.60                      & 37.89                      & -             & -             \\
			MattNet~\cite{MATT:}~\tiny{CVPR18}                  & mrcnn-resnet101                      &
			56.51                      & {62.37}                      & 51.70                      & {46.67}                      & {52.39}                      & {40.08}                      & \underline{47.64}         & \underline{48.61}         \\
			NMTree~\cite{liu2019learning}~\tiny{ICCV19}&mrcnn-resnet101   & 56.59 &\underline{63.02} &52.06&\underline{47.40} &\underline{53.01}& \underline{41.56}& 46.59 &47.88\\
			\hline
			MCN (ours)                      & vgg16                                & 57.33                      & 58.59                      & 57.23                      & 46.53                      & 48.68                      & 41.93                      &  46.95             & 47.20                  \\
			MCN (ours) & darknet53& \textbf{62.44}	&\textbf{64.20} &	\textbf{59.71}&	\textbf{50.62}&	\textbf{54.99}&	\textbf{44.69}
			& \textbf{49.22} & \textbf{49.40}\\ \hline
		\end{tabular}
	}
	\label{tab5}
	\vspace{-0.5em}
\end{table*}

\subsection{Evaluation Metrics}

For REC, we use the precision as the evaluation metric. When the \emph{Intersection-over-Union} (IoU) between the predicted bounding box and the ground truth is larger than 0.5, the prediction is correct. 

For RES, we use IoU and Acc@X to evaluate the model. The Acc@X metric measures the percentage of test images with an IoU score higher than the threshold X, while X higher than 0.5 is considered to be correct. 

In addition, we propose a \emph{Inconsistency Error} (IE) to measure the impact of the prediction conflict. The inconsistent results are considered to be the two types: 1) the results include wrong
REC result and correct RES result. 2) the results include
correct REC result and wrong RES result.

\footnotetext[1]{Scale denotes the resolution of the last feature map before prediction.}

\subsection{Implementation Details}
In terms of the visual backbone, we train MCN with Darknet53~\cite{redmon2018yolov3:} and Vgg16~\cite{simonyan2014very}.
Following the setting of MattNet~\cite{MATT:}, the backbones are pre-trained on MS-COCO~\cite{MSCOCO} while removing the images appeared in the val and test sets of three datasets. 
The images are resized to 416$\times$416 and the words in the expressions are initialized with GLOVE embeddings~\cite{pennington2014glove}. The dimension
of the GRU is set to 1,024. In terms of multimodal fusion, the project dimension  in Eq.~\ref{eq1} and Eq.~\ref{eq2} is 512. For the Soft-NLS,  we set $\alpha_{up}$ to 1.5 and set $\alpha_{dec}$ to 0.5.   We set the maximum sentence length of 15 for RefCOCO and RefCOCO+, and 20 for RefCOCOg. To binarize the prediction of RES, we set a threshold of 0.35. 

We use Adam~\cite{kingma2014adam} as the optimizer, and the batch size is set to 35. The initial learning rate is 0.001, which is multiplied by a decay factor of 0.1 at the 30th,  the 35th and 40th epochs. We take nearly a day to train our model for 45 epochs  on a single 1080Ti GPU. 
\begin{figure}[t]
	\centering
	\includegraphics[width=1\columnwidth,height=0.75\columnwidth]{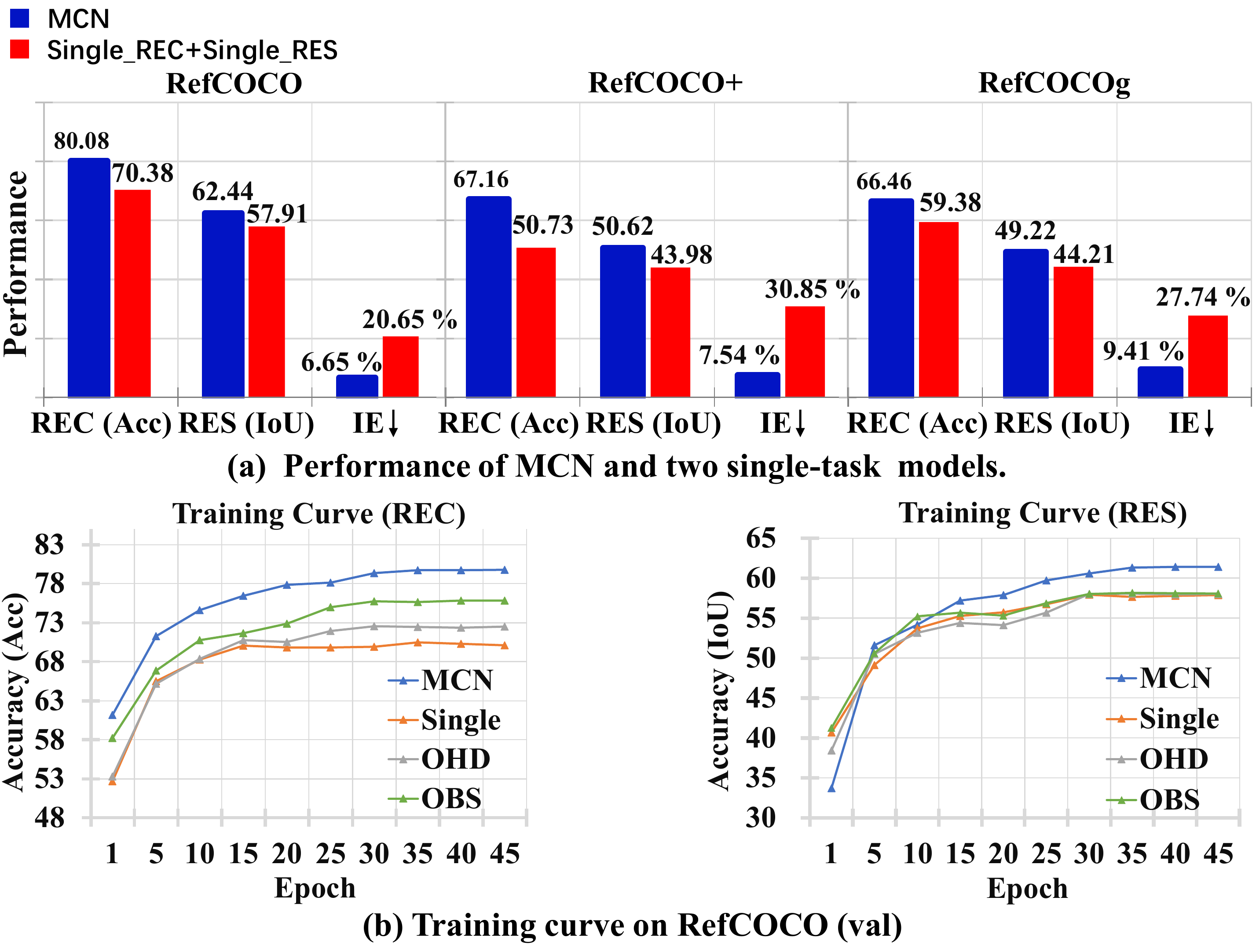}
	\caption{Comparisons of MCN and other  structures. (a) MCN significantly improves the performance of both two tasks on three datasets.
		(b) The learning speed of MCN is  superior to alternative structures.
		Here, all structures do not use the post-processing.}
	\label{quxian} 
	\vspace{-1em}
\end{figure}

\begin{figure*}[t]
	\centering
	\includegraphics[width=1.96\columnwidth,height=1.23\columnwidth]{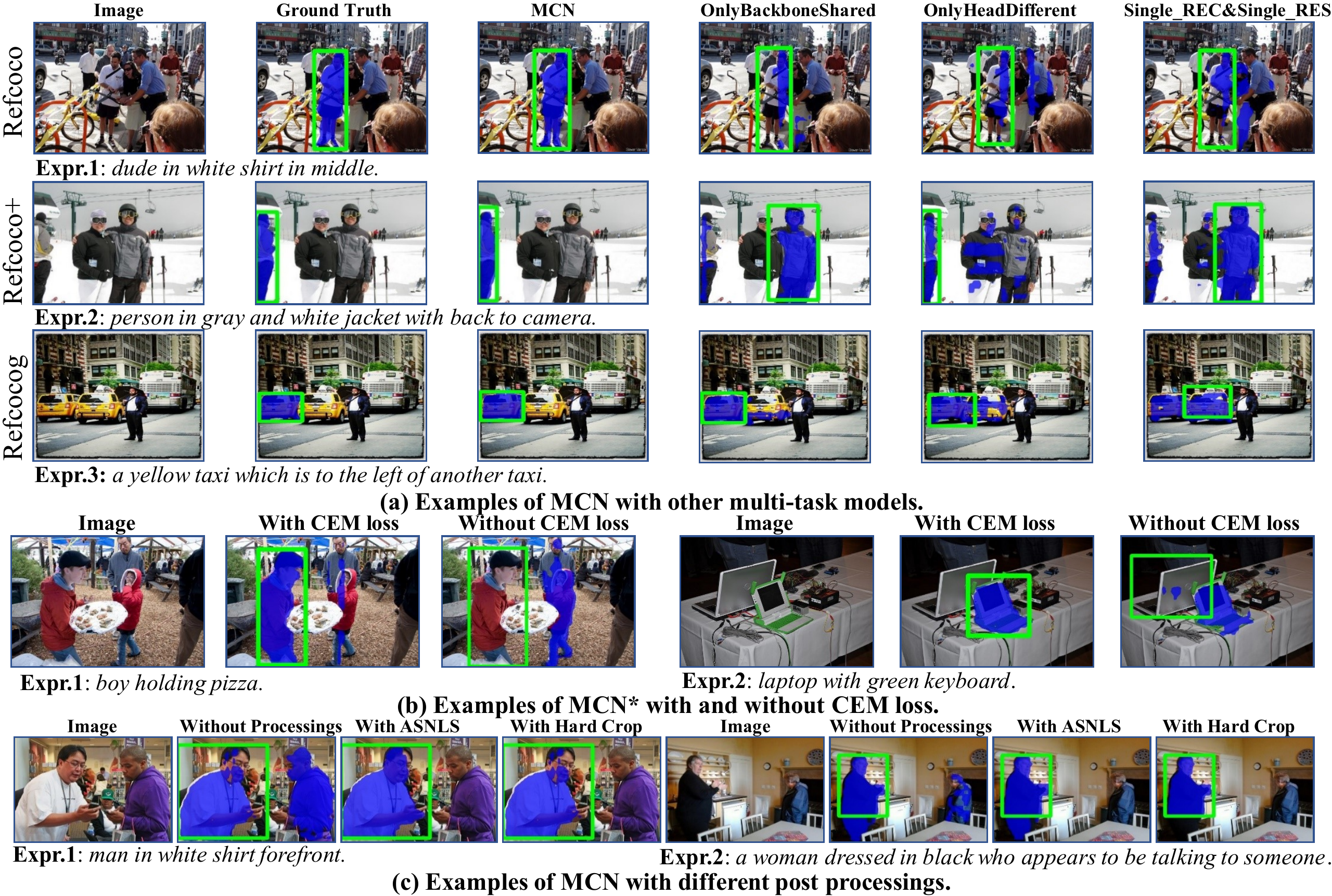} 
	\caption{Visualizations of the inference and prediction by the proposed MCN.  We compare the results of MCN with  three  multi-task networks in (a) and   compare the effects of our design in (b) and (c). * denotes that the post-processings is not used in these example.}
	\label{vis1}
\end{figure*}

\subsection{Experimental Results}
\subsubsection{Quantitative Analysis}
\textbf{Comparisons of different  network structures.
} 
We first evaluate the merit of the proposed multi-task collaborative framework, of which results are given in Tab.~\ref{tab1}.
In Tab.~\ref{tab1}, \emph{Single}\_\emph{REC} and \emph{Single\_RES} denote the single-task  setups.
\emph{OnlyHeadDifferent}~(OHD) and \emph{OnlyBackboneShared}~(OBS) are the other two types of multi-task frameworks. OHD denotes that the inference branches are also shared and only the heads are different, \emph{i.e.}, the regression layer for REC and the decoder for RES. 
In contrast, OBS denotes that the inference branches of two tasks are completely independent. 
From the first part of Tab.~\ref{tab1}, we observe that MCN  significantly benefits both  tasks.
Besides, we notice that the two tasks have different optimal settings about the scales of the multimodal tensors, \emph{i.e.}, $13 \times 13$ for REC and $52 \times 52$ for RES, suggesting the differences of two tasks. 
The second part of Tab.~\ref{tab1} shows that a completely independent or fully shared network can not maximize the advantage of the joint REC and RES learning, which subsequently validates the effectiveness of the collaborative connections built in MCN. Meanwhile,  as shown  in Fig.~\ref{quxian},  MCN demonstrates its benefits of collaborative multi-task training and  outperforms other single and multi-task models  by a large margin.


\textbf{Comparison of ASNLS and different post-processing methods.} 
We further evaluate  different processing methods,  and give the results in Tab.~\ref{tab2}.
From Tab.~\ref{tab2}, the first observation is that all the processing methods based on REC have a positive impact on both the RES performance and the IE score.  
But we also notice that the hard processing, \emph{i.e.}, RoI Crop~\cite{he2017mask,ren2017faster},  still reduces the performance of RES on some metrics, \emph{e.g.}, IoU and Acc@0.9, while our soft processing methods, \emph{i.e.} Soft-NLS and ASNLS, does not. 
This results greatly prove the robustness of our methods.
Meanwhile, we  observe that ASNLS can achieve more significant performance gains than Soft-NLS, which validates the effects of the adaptive factor design. 

\textbf{Ablation study.} Next, we validate different designs in MCN, of which results are given in Tab.~\ref{tab3}. 
From Tab.~\ref{tab3}, we can  observe  significant performance gains by each design of MCN, \emph{e.g.}, up to 7.04\% gains for REC and 14.84\% for RES.
We also notice that CEM  not only helps the model achieve distinct improvements on both the REC and the RES tasks, but also  effectively reduces the IE value, \emph{e.g.}, from 17.12\% to  13.51\%.
Similar advantages can also be witnessed in ASNLS.
Conclusively, these results confirm the merits of the collaborative framework, CEM and ASNLS again.

\textbf{Comparison with the State-of-the-arts.} 
Lastly, we compare MCN with the state-of-the-arts (SOTAs) on both REC and RES, of which results are given in Tab.~\ref{tab4} and Tab.~\ref{tab5}.
As shown in Tab.~\ref{tab4}, MCN  outperforms most existing methods in REC.
Even compared with the most advanced methods, like MattNet~\cite{MATT:}, MCN still achieves a comprehensive advantage and has distinct improvements on some splits, \emph{e.g.} +7.13\% on the testB split of RefCOCO and +2.80\% the val split of RefCOCO+.
In addition, MCN obviously merits in the processing speed to these multi-stage methods, \emph{e.g.}, 6 times faster than MattNet, which also suggests that the improvements by MCN are valuable. 
Meanwhile, MCN are signficantly better than the most advanced one-stage model, \emph{e.g.}, FAOA~\cite{yang2019fast}, which confirms the merit of the joint REC and RES learning again. 
In Tab.~\ref{tab5}, we  further observe that the performance leads of MCN  leads in  RES task is more distinct, which is up to +8.39\% on RefCOCO, +11.50\% on RefCOCO+ and +3.32\% on RefCOCOg.
As previously analyzed, such performance gains stem from the collaborative learning structure, CEM loss and   ASNLS, greatly confirming the designs of MCN.

\subsubsection{Qualitative Analysis}
To gain  deep insights into MCN, we  visualize its predictions in Fig.~\ref{vis1}.   The  comparisons between MCN and alternative structures are shown in Fig.~\ref{vis1}~(a).  From Fig.~\ref{vis1}~(a),  we can observe that the collaborative learning structure of MCN  significantly improves the results of both REC and RES.  Besides, MCN is able to predict high-quality boxes and masks for the referent in complex  backgrounds, which is often not possible  by alternative structures,  \emph{e.g.}, Expr.1. 
Fig.~\ref{vis1}~(b) displays the effect of the proposed CEM loss. 
Without it, the model tends to focus on different instances of similar semantics,  resulting the prediction conflicts of the REC and RES branches. 
With CEM, the two inference branches can have a similar focus with respect to the expression. 
Fig.~\ref{vis1}~(c) shows results of the model without and with different post-processing methods. 
From these examples, we can observe that the proposed ASNLS helps to preserve the integrity of an object, \emph{e.g.,} Exp.(2). It can be seen that the part of referent outside the bounding box is preserved by our ASNLS, while it will be naturally cropped  by the hard  methods, \emph{e.g.,}  ROI-Pooling~\cite{ren2017faster} and RoI-Align~\cite{he2017mask}. Conclusively, these visualized results reconfirm the effectiveness of the novel designs in MCN, \emph{i.e.}, the collaborative learning structure, CEM and ASNLS. 

\section{Conclusion}
In this paper, we propose a novel \emph{Multi-task Collaborative Network} (MCN) for the first attempt of joint REC and RES learning. 
MCN maximizes the collaborative learning advantages of REC and RES by using the properties of two tasks to benefit each other.
In addition, we  introduce  two  designs, \emph{i.e.}, \emph{Consistency Energy Maximization} (CEM) and \emph{Adaptive Soft Non-Located Suppression} (ASNLS),  to address a key issue in this multi-task setting \emph{i.e.}, the prediction conflict.
Experimental results on three  datasets not only witness the distinct performance gains over SOTAs of REC and RES,  but also prove that the  prediction conflict is well addressed. 

\paragraph{Acknowledgements.}
\footnotesize{This work is supported by the Nature Science Foundation of China (No.U1705262, No.61772443, No.61572410, No.61802324 and No.61702136), 
	National Key R\&D Program (No.2017YFC0113000, and No.2016YFB1001503), 
	and Nature Science Foundation of Fujian Province, China (No. 2017J01125 and No. 2018J01106).}

{\small
\bibliographystyle{ieee_fullname}
\bibliography{egbib}
}

\end{document}